# Determining the severity of Parkinson's disease in patients using a multi task neural network

María Teresa García-Ordás[1] · José Alberto Benítez-Andrades[2]  ·
Jose Aveleira-Mata[1] · José-Manuel Alija-Pérez[1] · Carmen Benavides[2]

## Abstract
Parkinson's disease is easy to diagnose when it is advanced, but it is very difficult to diagnose in its early stages. Early diagnosis is essential to be able to treat the symptoms. It impacts on daily activities and reduces the quality of life of both the patients and their families and it is also the second most prevalent neurodegenerative disorder after Alzheimer in people over the age of 60. Most current studies on the prediction of Parkinson's severity are carried out in advanced stages of the disease. In this work, the study analyzes a set of variables that can be easily extracted from voice analysis, making it a very non-intrusive technique. In this paper, a method based on different deep learning techniques is proposed with two purposes. On the one hand, to find out if a person has severe or non-severe Parkinson's disease, and on the other hand, to determine by means of regression techniques the degree of evolution of the disease in a given patient. The UPDRS (Unified Parkinson's Disease Rating Scale) has been used by taking into account both the motor and total labels, and the best results have been obtained using a mixed multi-layer perceptron (MLP) that classifies and regresses at the same time and the most important features of the data obtained are taken as input, using an autoencoder. A success rate of 99.15% has been achieved in the problem of predicting whether a person suffers from severe Parkinson's disease or non-severe Parkinson's disease. In the degree of disease involvement prediction problem case, a MSE (Mean Squared Error) of 0.15 has been obtained. Using a full deep learning pipeline for data preprocessing and classification has proven to be very promising in the field Parkinson's outperforming the state-of-the-art proposals.

María Teresa García-Ordás, Jose Aveleira-Mata and José-Manuel Alija-Pérez are contributed equally to this work.

José-Manuel Alija-Pérez
jmalip@unileon.es

Extended author information available on the last page of the article.

# 1 Introduction

Parkinson's disease comes about when certain neurons in the brain, gradually break down or die. This loss of neurons produce lower dopamine levels and it causes abnormal brain activity [2]. Parkinson's disease affects males more frequently than females [30] and the main symptoms include tremor or shaking, usually beginning in hands or fingers, rigid muscles, speech changes, loss of automatic movements, etc. [11, 12]. It impacts daily activities and reduces the quality of life concerning patients and their families and also it is the second most prevalent neurodegenerative disorder after Alzheimer in people over the age of 60 [27].

Parkinson's disease in advanced stages is easy to diagnose but in the early stages the diagnosis becomes truly difficult because the symptoms are unspecific as well as being different from one person to another. For example, tremor is said to be the most common symptom but it is not present in some patients who present different symptoms [23]. This difficulty in the early stages of Parkinson's disease is a powerful motivation for deep learning techniques to detect this disease early and be able to treat the symptoms [10, 16].

In this paper, we propose several deep learning configurations for two purposes. On the one hand, to determine if a person suffers from severe or non-severe PD (Parkinson's disease) taking into account the motor and total indicators of the UPDRS scale and, on the other hand, to predict the degree to which the disease has reached.

In this work, the dataset analyzes a set of variables extracted from voice analysis of patients, making it a very non-intrusive technique. One of the biggest limitations when using this dataset is the difficulty of accurately detecting total scale and motor scale using only this type of voice information. In our work we have taken advantage of the ability of neural networks to extract extra information from the data thanks to the use of autoencoders that simplify the data by eliminating possible noisy variables from the data. In addition, the multitask capacity of neural networks to make several different predictions at the same time has been successfully verified, even improving the results obtained by training specific networks for a task. At this point, a simultaneous classification and regression system has been proposed that allows the neural network to better determine the most relevant characteristics to predict with the greatest possible accuracy the level of Parkinson's according to the UPDRS metrics. According to the results obtained, the combination of autoencoders with classifiers and regressors, all in the same network architecture, has shown to achieve the best results to date.

The paper is organized as follows. Related work is presented in Section 2. The methodology of the different proposed techniques is detailed in Section 3. In Section 4, the experiments carried out are explained and the results are shown and compared in Section 5.

Finally, we conclude in Section 6.

# 2 Related work

The field of artificial intelligence has been for the last few years trying to help medicine in the diagnosis of diseases using images and data extracted from the patient. In the area of neurodegenerative diseases a lot of research has also been done with the main goal of being able to diagnose diseases early and be able to treat them as soon as possible. In the study of Alzheimer's disease, numerous studies have used deep learning techniques to diagnose the disease in patients. In [5], a study of multiple techniques was carried out, the best being the Bi-LSTM recurrent neural network. In addition, numerous studies have been done for this disease using brain imaging, many of them collected in Ebrahimighahnavieh's review [6].

Another widespread disease is Parkinson's disease. Numerous researchers have been trying for years to find ways to identify not only which people have it, but also how severe it is and how the disease will evolve for each patient. In the work done by Almeida in [1], multiple classifiers were used for PD detection and investigation of an approach without the use of audio fusion. The clinical impact of this approach is the possibility for the physician to use the application installed on a smartphone for PD detection.

Along the same lines, Lauraitis et al in [13] proposed an application capable of identifying in early stages degenerative diseases such as Huntington, Alzheimer or Parkinson with an accuracy of 86.4%.

More recently, Zhang et al in [31] proposed two methods based on time frequency analysis and deep learning using electroencephalogram images for parkinson detection.

Pahuja and Prasad in [21] proposed a novel method based on multi modal features and deep learning convolutional neural networks for Parkinson's disease prediction.

Taking into account recorded audios of voice conversations, Quan and his team developed an end-to-end model to detect Parkinson's using convolutional neural networks on the mel spectrogram of the audios with very promising results [25].

There are currently two main ways of measuring the progression of Parkinson's disease. These are the Hoehn and Yahr scale [8] and the Unified Parkinson's Disease Rating Scale (UPDRS) which is a widely applied index of disease severity [15].

The two UPDRS scores are total UPDRS and motor UPDRS scores. The Total UPDRS score includes 31 items contributing to three subscales: (I) Mentation, Behavior, and Mood; (II) Activities of Daily Living; and (III) Motor Examination [22]. The motor UPDRS evaluates the motor ability (III) of the patient on a scale of 0-108, and total UPDRS provides a higher range: 0-176.

Diagnosing Parkinson's disease is complex. It requires the evaluation of motor and non-motor symptoms and this evaluation is challenging and require the expertise and subjectivity of clinicians. For this reason, machine learning and deep learning techniques may assist physicians to facilitate the diagnosis process.

In recent years, techniques for the evaluation of the progression the disease and the prediction of risk have been researched and evaluated. Numerous works have examined the importance of features [4, 14, 28] in all fields. In the field at hand, knowing which features of the disease are most important is also immensely helpful in diagnosing it. In the work developed by Prashanth et al. [24], the authors proposed a method to estimate the stage (normal, early or moderate) and severity of Parkinson's disease using machine learning techniques such as ordinal logistic regression (OLR), support vector machine (SVM), AdaBoost and RUSBoost-based. Feature importance in PD (Parkinson's Disease), is also estimated using Random forests classifiers. They obtain 97.46% of accuracy and on the other hand, body bradykinesia, tremor, facial expression (hypomimia), constancy of rest tremor and handwriting (micrographia) were observed to be the most important features in PD.

In [26], an excellent prediction of motor outcome in PD patients was demonstrated by employing automated hyperparameter tuning and an optimal utilization of FSSAs (Feature subset selector algorithms) and predictor algorithms.

There are many other techniques for early detection using machine learning techniques. For example, logistic regression, random forests, boosted trees and support vector machines (SVM) are used in [24]. In their work, it is demonstrated that these techniques perform at a high accuracy and a large area under the ROC curve (95%) in classifying early PD from a healthy normal.

Nilashi et al. [19], use Incremental support vector machines to predict Total-UPDRS and Motor-UPDRS. The authors also use Non-linear iterative partial least squares for data dimensionality reduction and the accuracy measured by MAE for the Total-UPDRS and Motor-UPDRS were MAE 0.4656 and MAE 0.4967 respectively.

In view of the good results using machine learning techniques, many authors have opted to use hybrid methods. For example, in [18], the authors applied an expectation maximization clustering algorithm to cluster the dataset and after that, an ANFIS (adaptive neuro-fuzzy inference system) and SVR (Support vector regression) are used to predict the PD progression. Principal Component Analysis (PCA) was also used for the reduction in dimensionality. The results also indicated that the method which combines clustering, PCA and SVR was promising; obtaining an AUC of 0.99.

A similar procedure is followed in [20] in which Singular Value Decomposition (SVD) and the Adaptive Neuro-Fuzzy Inference System (ANFIS) are used to predict UPDRS scores. In this case, the authors obtained an RMSE = 0.687 in the motor class and RMSE= 0.677 in the total class of UPDRS.

PD progression is still currently being investigated. The capabilities of deep learning algorithms have not yet been completely utilized in the field of PD research and it is believed that by having an in-depth understanding of the data, it is possible to automate the Parkinson's Disease diagnosis to certain extent [9].

In fact, the techniques with which the more promising results in UPDRS scale have been obtained to date, are those based on deep learning. Grover et al. [11], proposed a deep neural network classification obtaining an accuracy of 62.73% for total classification and 81.67% for motor classification. Deep learning techniques are also used in [7] in whicha 1D convolutional neural network is proposed to analyze gait information and also to predict the severity of the disease with the Unified Parkinson's Disease Rating Scale (UPDRS). Their proposed algorithm achieved an accuracy of 98.7% in classifying gait data and an accuracy of 85.3% in PD severity prediction.

As we can see in previous works, the Unified Parkinson's Disease Rating Scale (UPDRS) assessment is the most used scale for tracking the progression of PD symptoms. The tracking process is invasive, time consuming and it requires the supervision of medical staff. In [3], Castelli et al. investigate the use of an innovative intelligent system based on genetic programming for the prediction of a UPDRS assessment, using only data derived from simple, self-administered and non-invasive speech tests.

## 3 Methodology

Many configurations of two well-known deep learning techniques have been evaluated to predict the progress of the PD pathology by taking two values into account, the motor and the previously defined total UPDRS scale. Different configurations have been carried out, both to predict the labels separately and to devise classifiers and regressors to predict the labels as a whole.

## 3.1 Dataset

The Parkinson's Telemonitoring Data Set [29] was created by Athanasios Tsanas and Max Little of the University of Oxford, in collaboration with 10 medical centers in the US and Intel Corporation who developed the telemonitoring device to record the speech signals.

The dataset is made uo of a range of biomedical voice measurements from 42 people with an early-stage of PD for six-months. The recordings were automatically captured by the patient at home so it is a non-invasive technique. Each row corresponds to one of 5,875 voice recordings of these individuals. The main aim of the data is to predict the motor and total UPDRS scores from all the extracted features.

An exploratory data analysis of the dataset has been done and the statistical information of each column can be seen in Table 1.

## 3.2 Data preprocessing: normalization

Generally, a dataset is made up of different features and since each characteristic is different, it also follows a different distribution. In these cases, it is very difficult for an artificial neural network to fit the data. To solve this problem, the MaxMin normalization has been

**Table 1** Statistic analysis of each feature in the evaluated dataset

|  | Mean | Std | 25% | 50% | 75% | Max |
| --- | --- | --- | --- | --- | --- | --- |
| Age | 64.8049 | 8.8215 | 58 | 65 | 72 | 85 |
| Sex | 0.317787 | 0.465656 | 0 | 0 | 1 | 1 |
| Test_time | 92.8637 | 53.4456 | 46.8475 | 91.5230 | 138.4450 | 215.4900 |
| Jitter% | 0.0061 | 0.0056 | 0.003580 | 0.004900 | 0.006800 | 0.099990 |
| Jitter(Abs) | 0.000044 | 0.000036 | 0.000022 | 0.000035 | 0.000053 | 0.000446 |
| RAP | 0.0029 | 0.003124 | 0.001580 | 0.002250 | 0.003290 | 0.057540 |
| PPQ5 | 0.0033 | 0.003732 | 0.001820 | 0.002490 | 0.003460 | 0.069560 |
| DDP | 0.0089 | 0.009371 | 0.004730 | 0.006750 | 0.009870 | 0.172630 |
| Shimmer | 0.0340 | 0.025835 | 0.019120 | 0.027510 | 0.039750 | 0.268630 |
| Shimmer(dB) | 0.3109 | 0.230254 | 0.175000 | 0.253000 | 0.365000 | 2.107.000 |
| APQ3 | 0.0171 | 0.013237 | 0.009280 | 0.013700 | 0.020575 | 0.162670 |
| APQ5 | 0.02014 | 0.016664 | 0.010790 | 0.015940 | 0.023755 | 0.167020 |
| APQ11 | 0.0274 | 0.019986 | 0.015665 | 0.022710 | 0.032715 | 0.275460 |
| DDA | 0.0514 | 0.039711 | 0.027830 | 0.041110 | 0.061735 | 0.488020 |
| NHR | 0.0321 | 0.059692 | 0.010955 | 0.018448 | 0.031463 | 0.748260 |
| HNR | 21.6794 | 4.2910 | 19.4060 | 21.9200 | 24.444 | 37.8750 |
| RPDE | 0.5414 | 0.100986 | 0.469785 | 0.542250 | 0.614045 | 0.966080 |
| DFA | 0.6532 | 0.070902 | 0.596180 | 0.643600 | 0.711335 | 0.865600 |
| PPE | 0.2195 | 0.091498 | 0.156340 | 0.205500 | 0.264490 | 0.731730 |

Mean, standard deviation, max value and the first, second and third quartile are shown used in this work. The following equation has been defined (1):

$$\hat{x} = \frac{x - x_{min}}{x_{max} - x_{min}} \quad (1)$$

The aim of MaxMin Normalization is fix the data (represented as $x$) in the range [0,1] taking into account the maximum and minimum data value.

### 3.3 Data classification and regression: multilayer perceptron (MLP)

Classification refers to predicting a label and regression refers to predicting a quantity.

In this work, different MLP configurations have been used with two purposes, firstly, for classification, with the aim of determining whether a patient has severe or non-severe PD, and secondly, for regression, to determine the degree of disease progression in a given patient.

To both purposes, an MLP architecture (with several modifications) has been used.

This network is made up of one input layer, one output layer and one or more hidden layers. In deep learning, more than one layers are usually used in order to learn the complex information of the input data. In an MLP, each neuron of the layer $n$ is fully connected with all the neurons of the layer $n+1$.

When the classification problem is binary, as in this case, the most common approach uses one output neuron with a sigmoid activation function (2) which represent the probability of the input (represented as $x$) to belong to the positive class.

$$f(x) = \frac{1}{1 + e^{-x}} \quad (2)$$

In the regression case, the activation function of the output layer is defined in equation as "relu" (3).

$$f(x) = \max(0, x) \quad (3)$$

### 3.4 Data reduction: autoencoder

An autoencoder is a neural network architecture which tries to learn a deep representation of the data by compressing the features. To do so, a symmetric architecture is build in which the number of input neurons is the same as the number of output neurons, and the middle layer has fewer neurons than the input and output layers. This middle layer is a bottleneck known as latent space. When we try to get the output of the net the input itself, the network is able to learn a deep and compressed representation of the data in the latent space. Of all of the layers before the bottleneck one is usually called encoder, and all the layers after it, make up the decoder.

An autoencoder has multiple applications such as dimensionality reduction, image compression, image denoising, feature extraction, anomaly detection, etc. In this case, a dimensionality reduction has been applied to our dataset. A vanilla representation of an autoencoder is shown in Fig. 1.

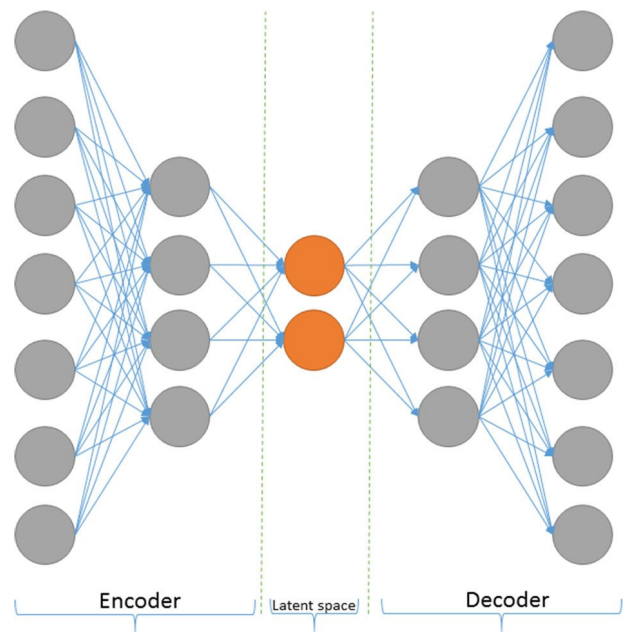

**Fig. 1** Vanilla autoencoder for dimensionality reduction

## 4 Experiments and results

### 4.1 Experimental setup

We have carried out two different experiments for all of the architectures: Classify samples (in severe and non-severe damage) and predict the UPDRS value. Firstly, experiments have been carried out to find out whether a person has severe or non-severe damage. Parkinson's Telemonitoring Data Set [29], was created to predict the motor and total UPDRS so for the purpose of classification, this scores are ranged following the work developed by Grover et al. [11]: If the total UPDRS value is above the score of 25, severe damage is considered. From 0 to 25, non-severe damage is considered. In the case of the motor score, if the score is above 20, it belongs to the severe class. From 0 to 20, it belongs to the non-severe class. All the following models were trained for classifying both motor and total scores.

Secondly, all the architectures have been tuned in order to return a numerical value to predict the UPDRS (motor and total) to find out at what stage the disease is found.

In all experiments, a grid search was performed to determine the optimal hyperparameters such as loss function, activation functions, batch size or learning rate.

#### 4.1.1 MLP

A basic MLP has been carried out with the aim of having a starting point to compare the obtained results. Our MLP architecture consists of four dense intermediate layers with 100, 200, 300 and 100 neurons respectively obtained experimentally, and two Dropout layers to avoid overfitting. In Table 2 an ablation study can be shown to demonstrate the optimal architecture. In all the cases the 4 layers configuration showed the best results.

**Table 2** Ablation study to determine the best layer configuration of MLP

| Hidden layers | Regression | | | Classification |
|---|---|---|---|---|
| | MSE | RMSE | MAE | Accuracy |
| y_motor | | | | |
| 1 | 519.2614 | 22.7873079 | 21.29 | 87.94 |
| 2 | 5.5454 | 2.3548673 | 1.6048 | 96.51 |
| 3 | 4.3517 | 2.08607287 | 1.3244 | 96.68 |
| **4** | **1.9603** | **1.40010714** | **0.8263** | **98.38** |
| 5 | 3.4661 | 1.86174649 | 1.1328 | 95.91 |
| y_total | | | | |
| 1 | 50.0422 | 7.07405117 | 5.4907 | 86.81 |
| 2 | 6.8737 | 2.62177421 | 1.8628 | 95.23 |
| 3 | 8.1247 | 2.85038594 | 1.9226 | 97.79 |
| **4** | **4.1276** | **2.03164958** | **1.2739** | **98.47** |
| 5 | 5.7784 | 2.40383028 | 1.5774 | 97.96 |

In the study, y_motor and y_total were taken into account for both classification and regression tasks. Best results were obtained in all the cases with the 4 layers configuration

The best results obtained from all the tested configurations for the 2 variables, y_motor and y_total, have been highlighted in bold

In all the experiments, 80% of the data has been used for training and the remaining 20% for testing the model. In both the classification and regression problems, the model has been fit for 1,000 epochs, and the batch size equals 20. A schema of these models can be shown in Fig. 2. From now, yellow boxes will represent Motor studies while green boxes will represent total score evaluation. In order to distinguish between classification and regression, we represent classification as a circle and regression as a rounded square.

### 4.1.2 MLP after autoencoder

The training data often contains unnecessary information that harms the network learning. For this reason, a feature reduction has been applied using an unsupervised autoencoder before carrying out the classification, thus eliminating non-relevant information.

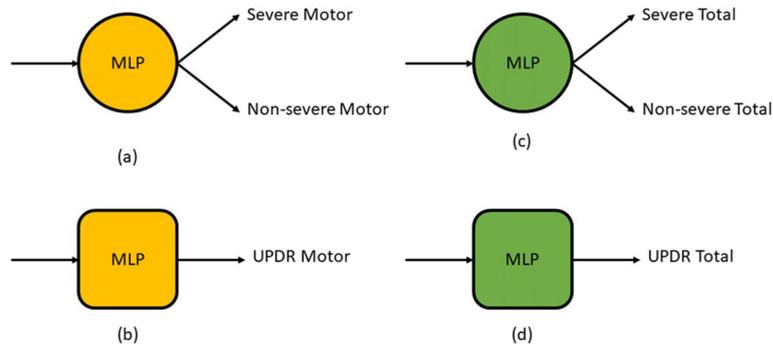

**Fig. 2** MLP models evaluated. (a) Classification of Motor severity. (b) Regression of UPDR Motor score. (c) Classification of Total severity. (d) Regression of UPDR total score

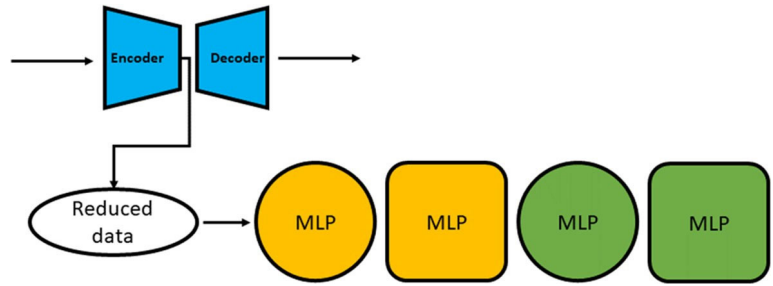

**Fig. 3** All MLP models trained after dimensionality reduction with an autoencoder

The autoencoder architecture is made up of one encoder with a dense layer of 200 neurons between the input layer and the latent space, and one decoder that maintains symmetry with the encoder.

The latent space has 10 neurons so the initial data is reduced to 10 features instead of the original number of features in the dataset (19) retaining only the most relevant information with the aim of improving the training in both classification and prediction steps.

Once we have our dataset reduced using the encoder of the trained autoencoder, these data have been used to train an MLP with the same architecture as in the previous section but by just modifying the input layer size. A schema of this proposal can be seen in Fig. 3.

### 4.1.3 Autoencoder with MLP

We have also combined the MLP classifier or regressor with the autoencoder jointly instead of first reducing the number of features and then classifying the new reduced data. In these experiments the classification or regression is carried out using the latent space. This architecture is very interesting because the encoder learns how to reduce the dimensionality of the data at the same time as it learns how to carry out a correct classification or prediction of the data. A diagram of the classification network can be seen in the Fig. 4.

### 4.1.4 Double task MLP

In the first experiments, we have trained one MLP for the total score and a different one for the motor score. As the UPDRS total score and the UPDRS motor score have a close relationship, we though that it would be interesting to use the same neural network for training both of them. So in this experiment, the total and motor scores (or classes) are used together as labels so we preserve information from both scores giving to the neural network a more powerful indicators during the training.

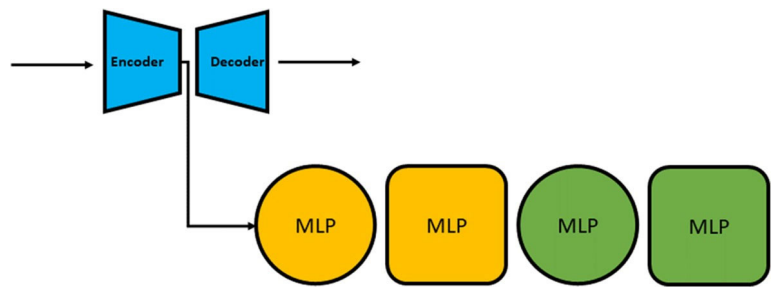

**Fig. 4** Architecture of the autoencoder with a classifier in the latent space

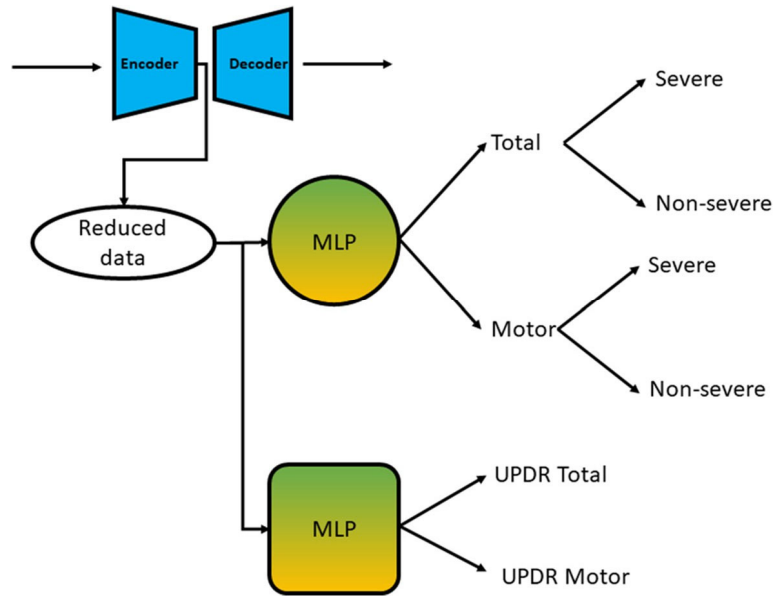

**Fig. 5** Architecture of the Double Task MLP. With this configuration, a classifier and a regressor have been trained for both scores at the same time

The MLP architecture is the same as that detailed above but the output layer contains two neurons, one per each score. The training for classification and regression purposes was done for 1,000 epochs and with a batch size of 20 (see Fig. 5).

### 4.1.5 Mixed classification and regression MLP

Finally, we proposed a combination of the classification net and that of the regression. In this case, an MLP has been proposed with two output layers and the reduced dataset obtained after training an autoencoder as input data. The first one with the aim of classifying the samples in severe or non-severe and the second one with the aim of predict the appropriate UPDRS. As we did in the basic MLP architecture, some dropout layers have been added between the dense layers in order to avoid overfitting. The entire architecture of this network for classifying severe or non-severe damage as well as predicting the score of the damage is shown in Fig. 6. In this case, two different nets have been trained, one for the total score and the other one for the motor score.

## 5 Results

### 5.1 Classification results

We have carried out all the experiments using an 80-20 train-test cross validation scheme. The results shown are the average of accuracy and MSE for all the repetitions. In Table 3 all the classification accuracy can be shown for Motor, Total and the average value of both scores. As we can see, the Double MLP just have an average metric due to the architecture of the network which merges both the Motor and the Total classifications into one net. The results show that the best performance are achieved with the mixed architecture which classify and predict the score at the same time. Not only is the average value the best, but so is the motor and total by themselves.

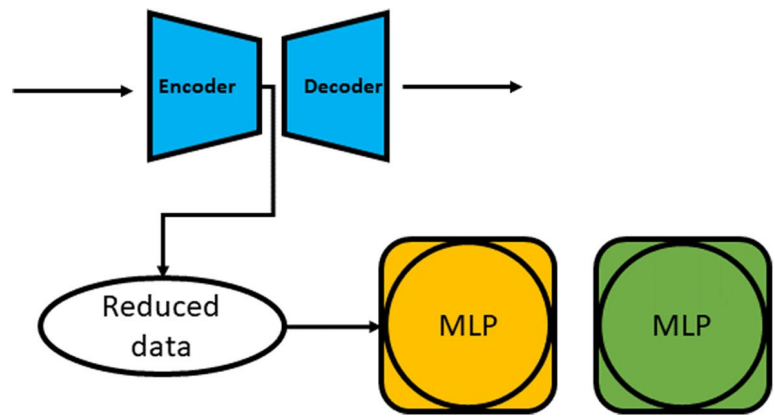

**Fig. 6** Scheme of the mixed model for regression and classification at the same time of motor (yellow) and total (green) UPDRS

For a better comparison of all the methods, in Fig. 7 we can see all the average values for all the architectures. As we said before, The mixed MLP outperforms all of the other methods obtained with the vanilla MLP with an increase in accuracy of 0.73%.

Furthermore, Our results have been compared with two other the state-of-the-art methods: ANFIS.SVR ([17]) and DNN ([11]) (see Fig. 8). As we can see, our experiments using deep neural networks using a reduced dataset obtained through an autoencoder, outperforms DNN by more than 37% in the average accuracy and ANFIS + SVR algorithm by 116.72%.

### 5.2 Prediction results

In Tables 4 and 5, the MSE, RMSE and MAE values obtained using the test subsets are shown for the Motor and Total UPDRS scores respectively. As we can see, the best results are also obtained using the mixed architecture followed by the double MLP net. The MLP

**Table 3** Accuracy results for all the architectures evaluated

| Net | Motor (%) | Total (%) | Average (%) |
| --- | --- | --- | --- |
| MLP | 98.38 | 98.47 | 98.43 |
| MLP after AE | 99.15 | 98.89 | 99.02 |
| Double MLP | | | 98.43 |
| MLP+AE | 98.98 | 98.89 | 98.94 |
| Mixed MLP | **99.32** | **98.98** | **99.15** |

In blue, the best result for each experiment

The best "accuracy" achieved from the models that have been tested in this research has been marked vanilla neural network achieved the worst result in both experiments, exactly the same as happened in the classification problem.

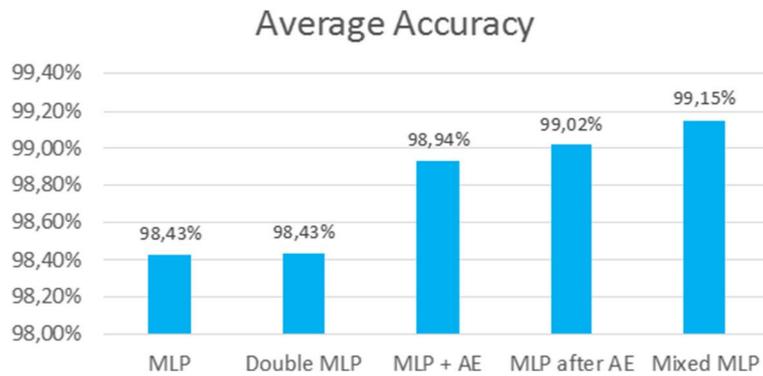

**Fig. 7** Motor and Total UPDRS average accuracy for all the experiments evaluated

As we did in the classification experiments, a comparison of all the average for the three regression metrics (MSE, RMSE and MAE) can be seen in Fig. 9. As we can see, MLP achieved a really bad performance in contrast with all the other methods which uses an autoencoder to reduce the number of features to 10. Mixed MLP improves the second-best architecture (Double MLP) by more than 60%. These results are very interesting for the medical community by evaluating the severity of Parkinson's disease in the patients automatically with a very low error.

Comparing these results with the state of the art shows the good performance of the proposed method also in regression. In [19], the authors achieved an average MAE of 0.4811 so our method outperforms it by more than 45%. More recently, in [20], the authors achieved

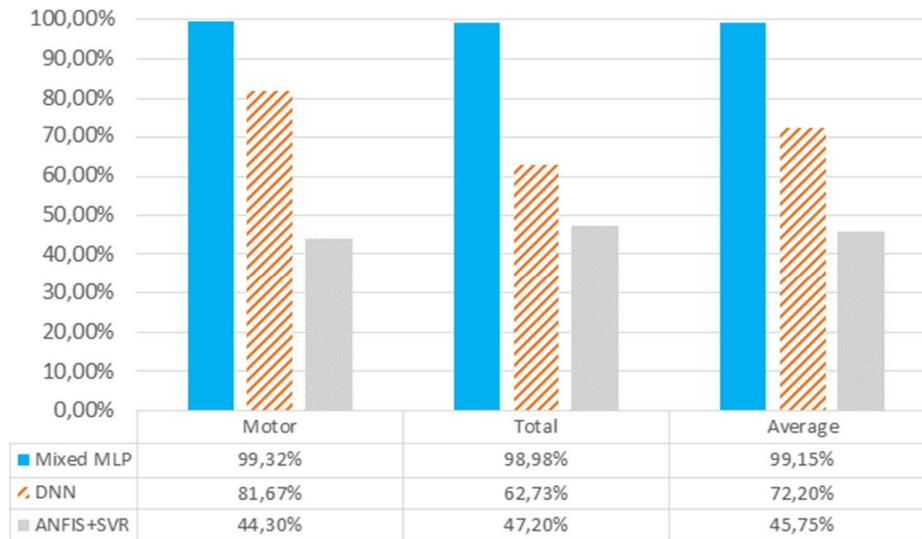

**Fig. 8** Our proposal classifier compared with state of the art

**Table 4** Regression results for motor score

| Net | MSE | RMSE | MAE |
| --- | --- | --- | --- |
| MLP | 1.9603 | 1.4 | 0.8263 |
| MLP after AE | 0.9616 | 0.9806 | 0.538 |
| MLP + AE | 0.8971 | 0.9471 | 0.6012 |
| Mixed MLP | 0.1399 | 0.374 | 0.2442 |

In blue, the best result for each experiment an average RMSE of 0.682 using SVD with ANFIS whereas our Mixed MLP obtains 0.40, increasing the performance by 41.35%.

## 6 Conclusions

In this paper, a new method for identifying the UPDRS (Unified Parkinson's Disease Rating Scale) in Parkinson's Disease patients has been proposed. Two different training tasks have been carried out: A classification for determining the severity of a Parkinson's Disease patient and a regression to predict its UPDRS. In order to extract the most relevant features of the dataset, an intelligent feature reduction was carried out by training an autoencoder. Experiments have demonstrated that both the classification and regression model achieved better results when they were feed with the reduced dataset. Furthermore, a new neural network architecture has been proposed which predicts the severity and the UPDRS for classification and regression at the same time. This allows the network to extract the optimal features for both evaluated tasks obtaining the best results until date with 99.15% of average accuracy in the classification task, outperforming the state of the art in more than 37%, and an average MSE of 0.1576 on the regression task which is 41.35% better than the most recent works.

The limitations of our proposal are mainly based on the nature of the input data. The set of data used is simply made up of data extracted from the voice of the patients, which makes our method a non-intrusive and easy-to-apply solution, but limited in terms of accuracy to the data used. Even so, the results obtained demonstrate the great precision of the model when it comes to identifying early cases of Parkinson's and its ability to determine the severity of the disease, both motor and general. In the future, these characteristics could be combined with the characteristics extracted from encephalogram images or from the voice

**Table 5** Regression results for total score

| Net | MSE | RMSE | MAE |
| --- | --- | --- | --- |
| MLP | 4.1276 | 2.031 | 1.2739 |
| MLP after AE | 1.0412 | 1.02 | 0.607 |
| MLP + AE | 0.9196 | 0.9589 | 0.8697 |
| Mixed MLP | 0.1753 | 0.4186 | 0.2857 |

In blue, the best result for each experiment recording wave itself, modifying the architecture so that it accepts this type of input data using convolutional layers.

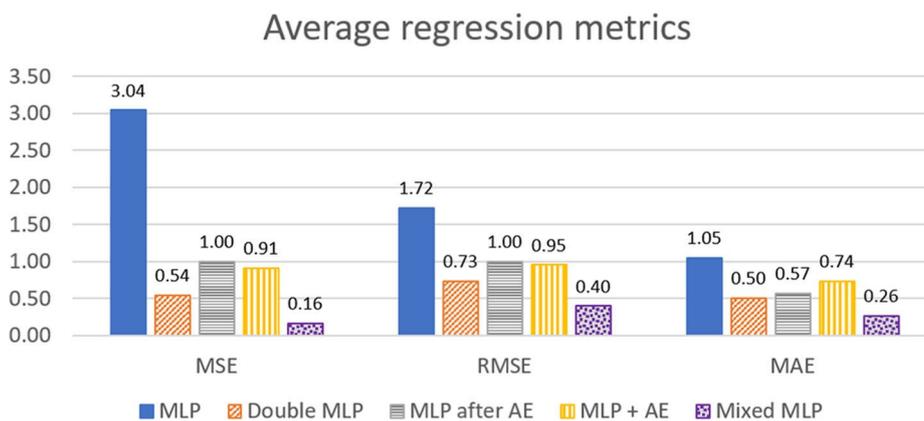

**Fig. 9** Motor and Total UPDRS average MSE, RMSE and MAE for all the experiments evaluated

In conclusion, the achieved results demonstrate the capability of deep learning to determine the severity of Parkinson's disease with a high degree of reliability.

**Author Contributions** **María Teresa García-Ordás**: Conceptualization, Data curation, Methodology, Software, Visualization, Validation, Writing- Original draft preparation. **José Alberto Benítez-Andrades**: Data curation, Methodology, Software, Visualization, Validation, Writing- Reviewing and Editing. **José-Manuel Alija-Pérez**: Data curation, Writing- Original draft preparation. **Jose Aveleira-Mata**: Conceptualization, Supervision, Writing- Reviewing and Editing. **Carmen Benavides**: Conceptualization, Supervision, Writing- Reviewing and Editing.

**Funding** Open Access funding provided thanks to the CRUE-CSIC agreement with Springer Nature. This research was funded by the Junta de Castilla y León grant number LE078G18.

**Data Availability** All the data used in the experiments, are available in Kaggle https://www.kaggle.com/rishidamarla/parkinsons-telemonitoring-data

## Declarations

**Conflict of Interests/Competing interests** The authors declare that they have no known competing financial interests or personal relationships that could have appeared to influence the work reported in this paper.

## Affiliations

**María Teresa García-Ordás[1]** · **José Alberto Benítez-Andrades[2]**  · **Jose Aveleira-Mata[1]** · **José-Manuel Alija-Pérez[1]** · **Carmen Benavides[2]**

María Teresa García-Ordás
mgaro@unileon.es

Jose´ Alberto Ben´ıtez-Andrades
jbena@unileon.es

Jose Aveleira-Mata
jose.aveleira@unileon.es

Carmen Benavides
carmen.benavides@unileon.es

[1] SECOMUCI Research Group, Escuela de Ingenierías Industrial e Informática, Universidad de León, Campus of Vegazana s/n, León, 24071, León, Spain

[2] SALBIS Research Group, Department of Electric, Systems and Automatics Engineering, Universidad de León, Campus of Vegazana s/n, León, 24071, León, Spain